# Go game formal revealing by Ising model

Matías Alvarado[a], Arturo Yee[b], and Carlos Villarreal[c]

[a] *Center of Research and Advanced Studies – IPN, Department of Computer Science, Av. Instituto Politécnico Nacional 2508, San Pedro Zacatenco, 07360, Ciudad de México.*
[b] *Universidad Autónoma de Sinaloa, Facultad de Informática Culiacán, Sinaloa, México.*
[c] *National Autonomous University of Mexico (UNAM), Institute of Physics, Circuito de la Investigación Científica, Ciudad Universitaria, 04510, Ciudad de México.*

**Abstract**

Go gaming is a struggle for territory control between rival, black and white, stones on a board. We model the Go dynamics in a game by means of the Ising model whose interaction coefficients reflect essential rules and tactics employed in Go to build long-term strategies. At any step of the game, the energy functional of the model provides the control degree (strength) of a player over the board. A close fit between predictions of the model with actual games is obtained.

## 1. Introduction

Mathematical modeling and algorithmic setting of Go game is a meaningful problem in the state of the art of sciences. The complex interaction among elementary black and white stones over a squared board in Go gaming, looks-like similar to the modeling of complex interaction from simple elements in major nature [24, 25] and social phenomena [26, 30]. Formal analysis of Go game is core in advances in computer science likewise the analysis of Chess was during the 20th century [1]. We use the Ising model [24] classical tool for modeling dynamic changes in complex interaction to fundament the algorithm to quantify the cooperation strength among allied stones or tension against the adversary stones struggling in a Go game beat. At some moments during a Go match, a phase-transition-like process corresponds to strong preeminence of blacks over white or conversely. The ever intricate Go interaction and the measure of each gamer' strength at any game step, as the major challenge for Go automation, it is fine model with the Ising Hamiltonian.

1.1 Go game

Go is a 2 players, zero-sum and complete information game, black versus white stones that official goban board is a 19 x 19 grid [2]. By turn, each player places one black/white stone on an empty board cross-point position; black plays first then white and so on. White receives a compensation komi by playing the second turn [3]. Goal of a GO successful game is to get the most of the board territory by means of GO tactics of invasion, reduction, nets, ladders and connections. One stone's liberty is a contiguous empty board cross-point in the vertical or horizontal direction. Stone allocation in an empty board neighborhood is an invasion, and if adversarial places one stone close to invasion it does a reduction. Same color stones do a net over adversarial stones by surrounding them, and do a ladder by surrounding and leaving a sole liberty, Atari, to adversarial stones. Connection of ally stones is by placing one same color stone between them. Same color stones joined in horizontal or vertical line form up one indivisible compound stone, hence single or compound stones are struggling for achieving territory control. A stone placed on board is captured by adversary reducing their liberties to zero then removed. Principal concepts for GO are being alive or dead. A stone is alive if cannot be captured and is dead if cannot avoid be captured. Placement of stone being directly captured is suicide that is not allowed. Go strategies follow sequences of tactics aimed for the most board area control. The game ends when both players pass turn. The score is computed based on both board territory occupied and the number of adversarial single stones captured. The winner has the largest territorial control and the largest number of captures, as the most usual criteria.

The hardest task for Go gaming is to evaluate the control board area and the dominion status of a player at a given stage of the game, for humans or computer players. Fig. 1 shows the flow diagram for computer Go gaming, which disarming simplicity not avoids a complex combinatorial process to attain efficient strategies [11, 12].

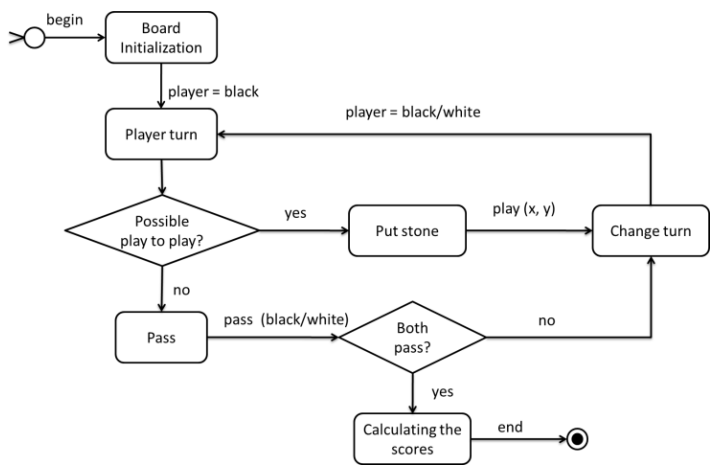

**Fig. 1** Go gaming flow diagram

1.2 Computer Go

To overcome the above mentioned Go gaming complexity was a major challenge for human Go players, as well as by last two decades for computer Go scientists and developers [2], [13]. Categorical computer AlphaGo's triumph of 4/5 over Lee Sedol, one of the best world Go player, was a meaningful triumph of computational intelligence by 2016 [42]. Top preeminence of AlphaGo is the 60 - 0 simultaneous triumphs over the best human Go players by February 2017 [14], [15, 16].

A Go gaming state or Go state is a board configuration given by the combination of black-white-empty board Go gaming positions. The Go state space cardinality is $3^{19 \times 19} \sim 10^{172}$. The game tree records the different game decisions the players can try during a game from the start to the end, so records the different paths between the got states. One path from start to end state registers the sequence of moves in a match. Go game tree cardinality is by $10^a$, $a = 10^{172}$. As a result, the automation of successful Go tactics and strategies, translated in efficient paths to win a match, is hugely complex. In average, the branching factor for Go ranges 200 - 300 possible moves at each player's turn, while 35 - 40 moves for Chess, which state space and game tree cardinality is $10^{50}$ and by $10^{123}$ [11]; respectively $10^{18}$ and $10^{54}$ for Checkers.

Hard task in Go automation as for human Goers as well, is to estimate the potential to strength territory dominance for a given try, that implies classify the best sequence of states pick from the enormous number of Go states to decide next advantageous Go move [3, 17, 18]. Computer Go [10] uses heuristic-search [15, 16], machine learning [17] and pattern recognition techniques to identify eyes, ladders and nets [18], [19], [9]. Monte Carlo Tree Search (MCTS) was extensively used for simulation-based search algorithms [15, 20, 21] through the record of moves of previous games, and given a Go state, from thousands or millions simulations the best average is applied to ponder the next movement [15, 20], that is highly time consuming. The no apply of a-priori-knowledge to identify tactics or strategies is the major drawback of MCTS usage in computer Go [22, 23] [23]. In computer (cost) complexity to solve a problem [39], *time* refers to the number of execution steps an algorithm used to, and *space* refers to the memory amount used to. Go gaming complexity is EXPTIME-complete [40], and more precisely PSPACE-complete [41].

1.3 Recognition challenge

The recognition and discrimination of meaningful perceptual stimuli presupposes the active formation of stable perceptual elements to be recognized and discriminated. First, consider the spontaneous grouping of stones of the same color which occurs during visualization of a GO board. The stones are organized into distinct groups, clusters, or armies even though they may be sparsely scattered about or somewhat intermingled. Grouping is usually the result of proximity of stones of the same color or the predominance of stones of one color in an area, but can be affected by other characteristics of the total board situation. Closely related in grouping is segmentation, which is also discussed in Kohler. The area subtended by the board is divided into black and white territories, each of which maintains its own integrity in the visual field. These segments are a measure of the territory which is controlled by either side, hence are an important factor in the assessment of a GO board. Another example is the formation of "spheres of influence" about a stone or group of stones. Influence is not an inherent property of stones, but appears to be induced in them by our process of perception. Yet they are a crude measure of the potential of a stone or army of stones for controlling territory on the board. The spontaneous image formed by the visualization of a GO board appears to be a complicated assemblage of perceptual units and subunits. For example, the stones themselves have their own perceptual identity while at the same time they are parts of chains or groups of stones. This report will describe a simulation model for visual organization. It will use transformations which create information corresponding to the perceptual features discussed above, storing them in a computer internal representation.

Our point is the use of the Ising model and Hamiltonian to precise characterize the intense interaction in Go gaming, as for interaction in electromagnetism and thermodynamic is made. In Section 2 the CFG (Common Fate Graph), the Ising model and the energy function to estimate the synergy strength of each set of ally stones in a Go state are introduced. In Section 3 experiments and the analytical comparison of results follow; Section 4 is for Discussion and Section 5 for Conclusions.

**Table 1**.

**2. Go gaming synergy**

In our proposal a Go state is represented by a CFG [6], being essential to stand back efficient automation. CFG represents a Go state by means of principal and secondary nodes, see Fig 2: each Go stone in the state is a CFG principal node labeled with the number of single stones that compose it, and each stone's liberty is a CFG secondary node. Go state representation by CFG it neatly comprises the each stone' size, the linked relationship with allies, with adversaries, and with liberties around each of the stones. Using CFG, Go sequence of moves, so the deployment of tactics and strategies during a Go game is easy logged, thus follow up the evolution of game interaction –depicted in a lattice graph–, formalized by the Ising energy function and the algorithms do quantify the force of atoms or molecules of black and white stones regarding to the relative position among allies and adversaries on board.

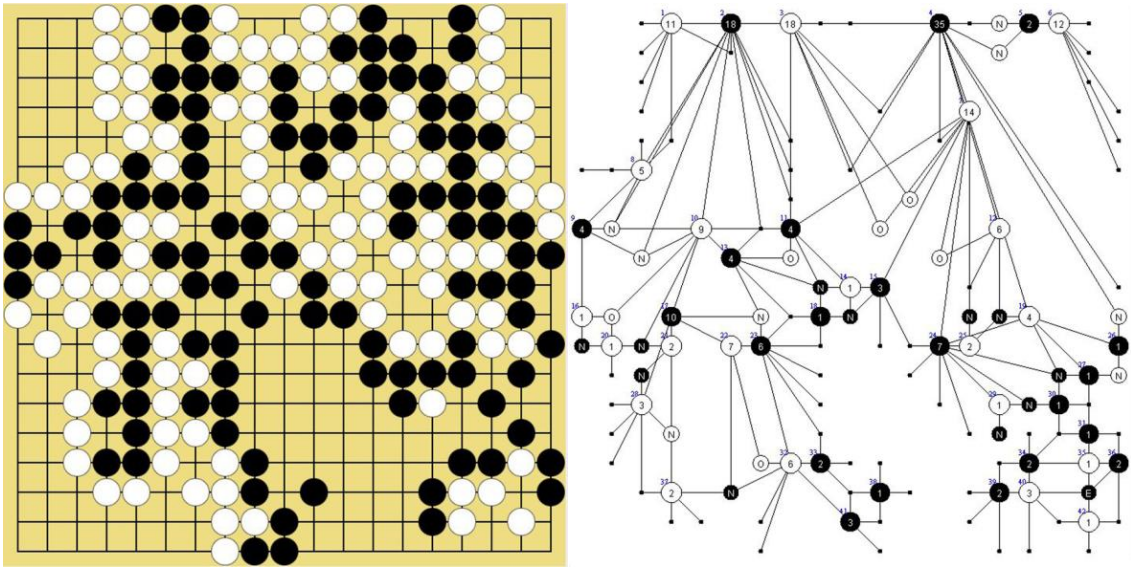

**Fig. 2** White stone in the left superior corner is 11 single stones, 6 liberties, 1 shared with the below white 5 single stones, and one black liberty shared with at right black 18 single stones.

2.1 Ising model and Hamiltonian

Ising model describes magnetic properties of materials from the interactions of constituent atomic spins, as elementary magnetic moments which possess a dichotomy behavior that point randomly in the up or down directions; variables $x_i$ represent the ith-spin state and may acquire dichotomy value 1 or -1. Each spin interacts with neighboring spins or with external fields that tend to align them in the applied direction, and depending on the strength of interactions, whole system get phase transitions among emergent spin clusters domains, percolate through the whole system, or fill out complete regions of the material. The spins are arranged in a n-dimensional lattice, which *energy interaction* is described, restricted to n=2, by the Hamiltonian Eq. (1):

$$H = -\sum_{ij} w_{ij} x_i x_j - \mu \sum_i h_i x_i \qquad (1)$$

$w_{ij}$ sets for interaction between spin $i$ and $j$, $\mu$ the magnitude of an external magnetic field, and $h_i$ the magnetic field contribution at site $i$; for a homogeneous external field, $h_i = 1$.

Phase transition by heat transfer it transforms a thermodynamic system from one state of matter to another, usually between solid, liquid and gaseous states of matter [24]: a liquid that become gas upon heating to the boiling point results in an abrupt change in volume. Hence, properties of the matter suddenly change as a result of the change of condition on temperature, pressure or others. Phase transitions are common in nature phenomena [25, 32, 36] in Biology, Physics or Chemistry, and is applied in cutting edge top technologies [31, 33, 38].

2.2 The Go game energy function

Our definition of energy function stands back algorithms to compute the power of stone patterns in a Go state, this way account board dominance. We use specific 2 dimensional Ising model for displaying phase transition in Go board. The energy function uses CFG representation of Go states to captures the interaction among allied stone, versus adversaries, as well as with respect to liberties involved in. Associated to Ising Hamiltonian in Eq. (1), the Go energy function should embrace parameters to deal with:

- The numbers of atomic (single) stones in a molecular (compound) stone.
- The number of eyes a stone is involving.
- The tactic pattern the stone is making.
- What's the strength of ally stones that do synergy among it.
- What's the strength of adversary stones that adversarial fight.

To get this goal, we propose the quantitative description of stone $i$ by means of the elements involved in Eq. (2):

$$x_i = c_i(n_i + r_{eye}^{k_i}). \qquad (2)$$

$n_i$ sets the number of single stones, $r_{eye}$ is the constant to represent the occurrence of an eye, $r_{eye} > 1$, or $r_{eye} = 0$ if no eye; $k_i$ sets the number of eyes in stone $i$, and $c_i$ is the stone color, 1 for white, and -1 for black. Hence, $r_{eye}^{k_i}$ quantifies the impact of the number of eyes in $i$, and $k_i = 2$ says that could never be captured. If no eye $x_i$ just indicates the $i$ size and color.

In Hamiltonian of Ising model for Go parameter, $w_{ij}$ sets the ratio of union or repulsion between each pair $i, j$ of single or compound stones, so $w_{ij}$ encompasses tensions alongside paths joining stone $i$ to $j$, being affected by the presence and strength of adversary stones that may impede the $i - j$ connection; or, as well as, by the presence of allied stones that result in mutual strengthen. Up to rules and tactics in Go gaming the feature interaction among stones is assessed by means of next Eq. (3):

$$w_{ij} = \sum_s r_t x_s^{(ij)} \qquad (3)$$

$x_s^{(ij)}$ describes each stone $s$ lying between $i$ and $j$, that makes a tactic pattern; $r_t$ sets a quantify of *a-priori* known power of the pattern $t$: eye ($r_{eye}$), net ($r_{net}$), ladder ($r_{lad}$), invasion ($r_{inv}$), reduction ($r_{red}$). Pattern parameters fit a total order $>$ induced by *a-priori* knowledge of Go tactics power, by an averaging procedure from real matches between top level players –estimation open to analysis and precisions: we say that an eye tactic has top power, followed by a net, a ladder, an invasion and a reduction. Thus, $r_{eye} > r_{net} > r_{lad} > r_{inv} > r_{red} > r_{sl}$. Single liberty parameter value is $r_{sl} = 1$.

In the Go game perspective, the first term of Hamiltonian in Eq. (1) accounts the interaction of collaboration among patterns of same color stones, or the fight against adversaries; for second term, the particular external field $h_i$ adds the number of liberties the stone $i$ has. Henceforth, given any Go state, by definitions in equation (2) and (3) used in equation (1) we quantify the every stone's power and, on the base of each of them, the interaction strength among ally and/or adversarial stones: what's the contribution of each eye, ladder or net pattern; or by invasion, reduction or connection tactics.

2.3 Phase Transition

On the other hand, the evolution leading to territory control as a result of phase-transition-like process, like parallel version of changes of matter by heat transmission in nature phenomena, is follow up. The *heated* stones sequentially placed as a Go move it eventually change the state board abruptly, similar to matter changes. This sequence of moves (states) yields to a Go *phase-transition* process that it brings sudden board area dominance. In Fig 3 the sequential placement of red-black-flag stones makes the override over white in this board area, alike sequential local phase transition.

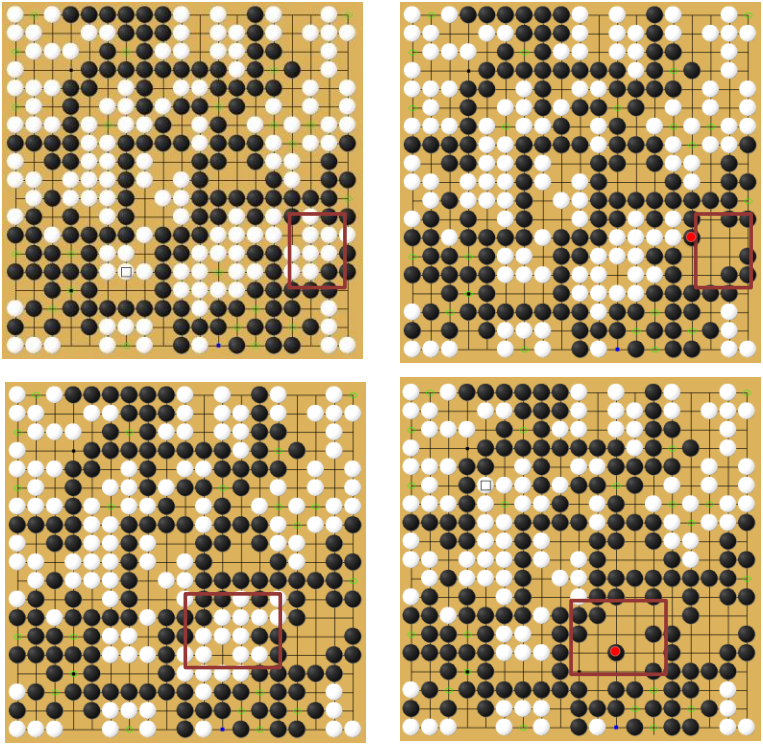

**Fig. 3** Go phase-transition by placement of black stones.

## 3. Experiments

We use Go games' information in http://www.go4go.net/ to evaluate this proposal. Results of experiments show the worth using Ising Hamiltonian to compute the synergy strength of Go state patterns during games. Actually, close similar results to the ones in top official tournaments [5], including the AlphaGo games against human Go players, are obtained.

The standard format of smart game files (SGF) containing the decision game trees of games [4] is CFG translated to quantify any Go state by the energy function.

3.1 Human Go games

In experiments the energy function is applied to quantify the patterns' strength in every Go state during a match. The strength of black and white stones patterns is valued to pinpoint what is better placed on board at each state in the sequence. After each move comparison between black versus white patterns' strength by energy functions follows. In next figures blue line is for blacks and red line is for whites.

Classical Go games in Fig 4 and Fig 5, well known games are the input to simulator. In 1961-09-13,14, both 9p, Sugiuchi Masao, play black stones versus Fujisawa Hosai, the winner that plays white stones.

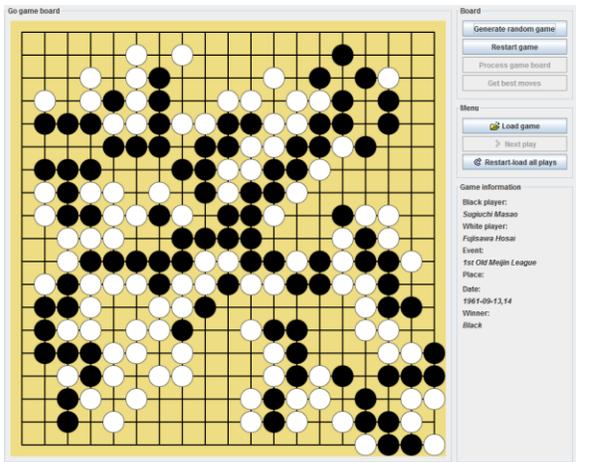

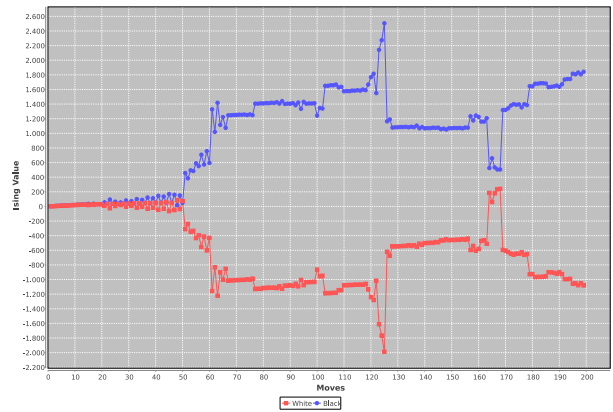

In Fig 5, the game in 2003-04-23 , between Yi Se Tol played black stones and beat to white stones player Hong Chang Sik.

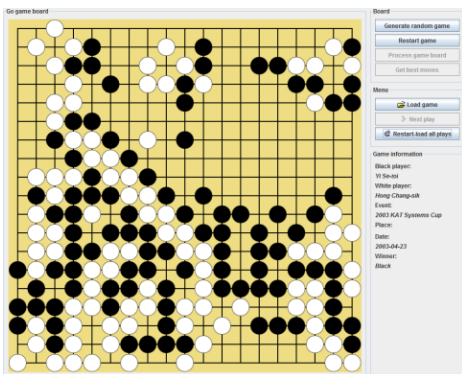

10/19/2017 2:14 PM

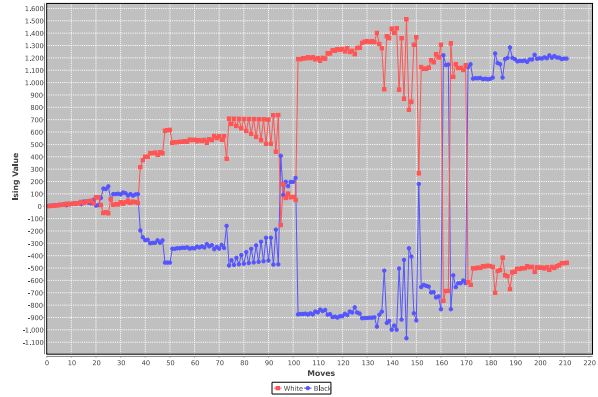

In Fig 6 we illustrate the dynamic of the match played by **Murakawa** with black stones versus **Cho Chikun** with whites in the 39th Japanese Kisei. Since the start and during one hundred movements, according to the valuation of the energy function, almost equal strength of black and white stones happened.

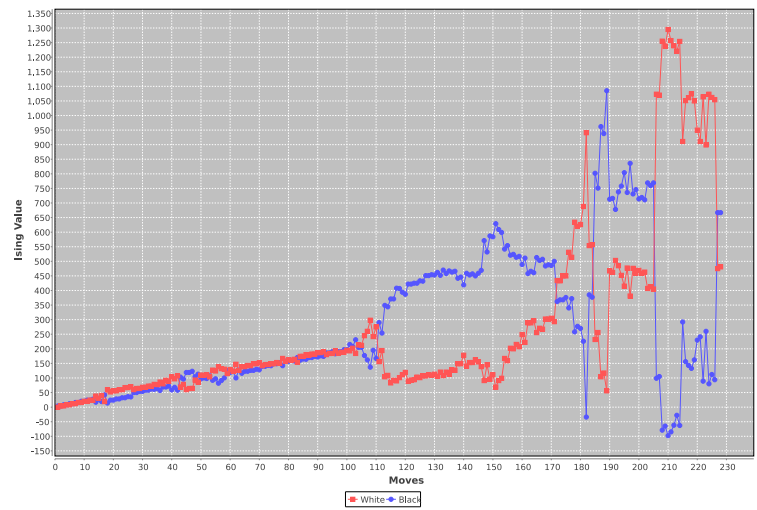

**Fig. 6** During moves 100 – 170 the stones patterns' strength values are separated, black strength favored, the separation becomes bigger during moves 171 – 230 and white strength favored. At the end, the strength of black stones is 667 and is 481 for white, as our energy function estimates. This match score reported from human organization is similar win for black as well.

In Fig 7 graph shows the comparison of strength of Go patterns in match **Lee Changho** with black stones versus **Ryu Suhang 3p** with whites, in Korean League 2014. According to energy function calculus the strength of black stones is 424 and for the white player is -51, likewise result reported. The black-whites balanced dominance breaks by the placement of a single stone that leads to capture a lot of adversarial stones and mostly control the formerly shared area.

In described experiments the final score using energy function is according to the report from each tournament's organizers of top qualified Go players. We made dozens of similar experiments using the Ising model energy function to evaluate the interaction and stones patterns strength, from initial to final states in Go matches that is reported in http://www.go4go.net/. In 19 off 30 matches we got correct result, 8 off 30 we got minor error than 5%, and we got a bit non-correct result error, in a range of 6% - 10%, close to the correct match score. Please, see related material in http://delta.cs.cinvestav.mx/~matias/Teoria_Juegos/Go/EnergyFunction.

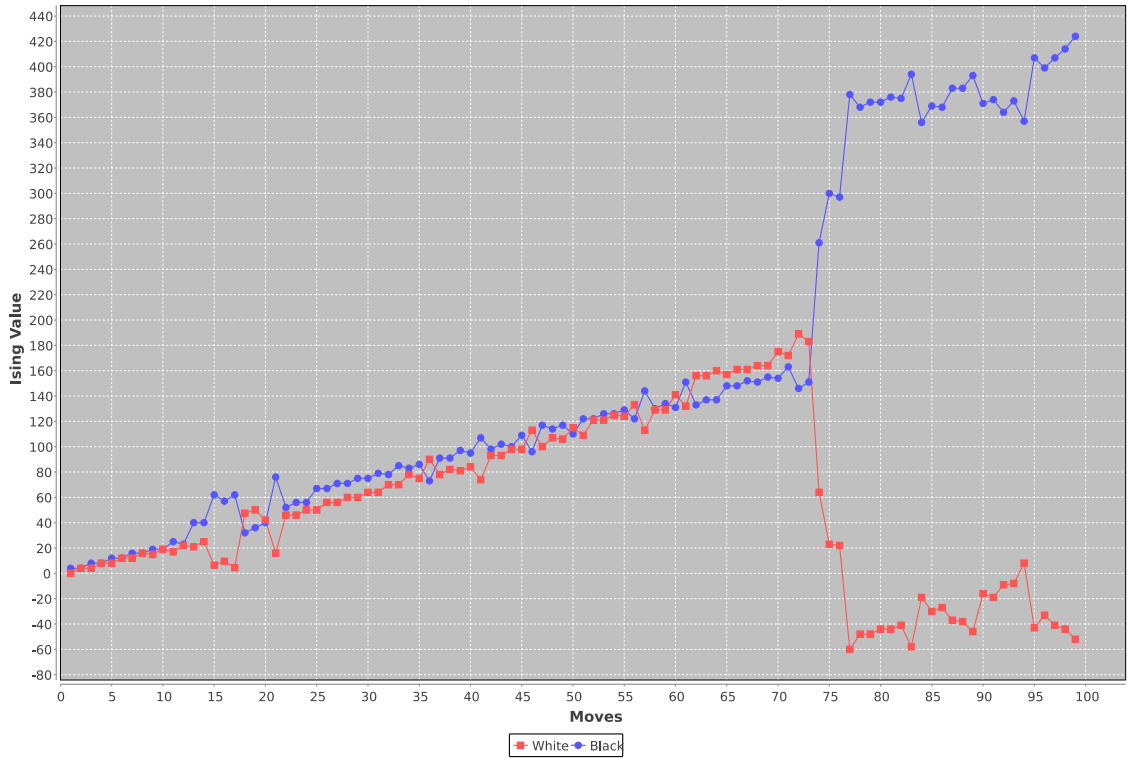

**Fig. 7** A close fight happens in around 70 off 100 movements, so each stones pattern strength were similar there. At move-state 73 a phase transition occurs and strength for blacks becomes each late state greater than the strength for whites, until the game end.

## 3.2 AlphaGo versus humans players

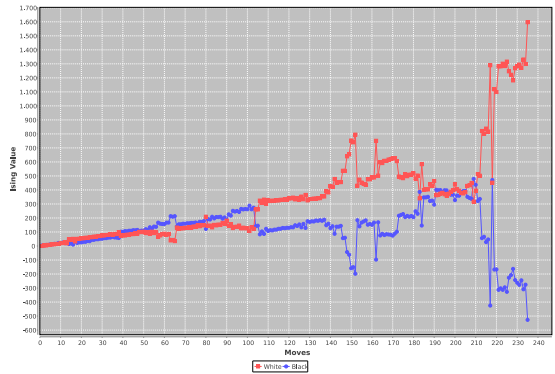

(a) Gu Li vs. AlphaGo 2017

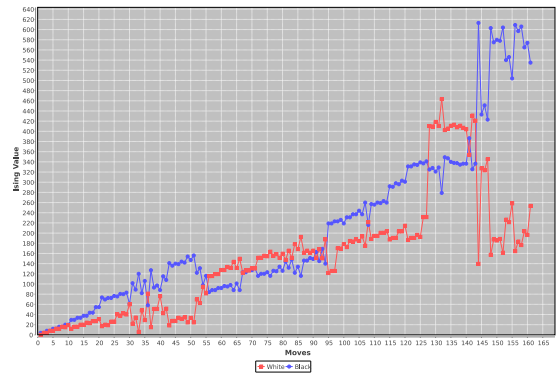

(b) AlphaGo vs. Zhou Ruiyang 2017

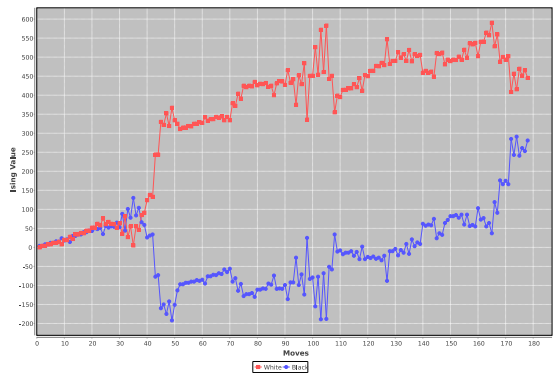

(c) Chang Hao vs. AlphaGo 2017

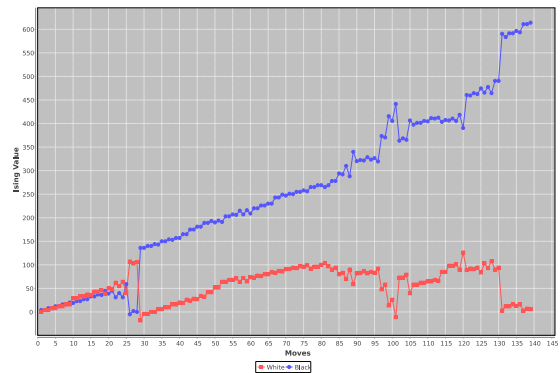

(d) AlphaGo vs. Shin Jinseo 2017

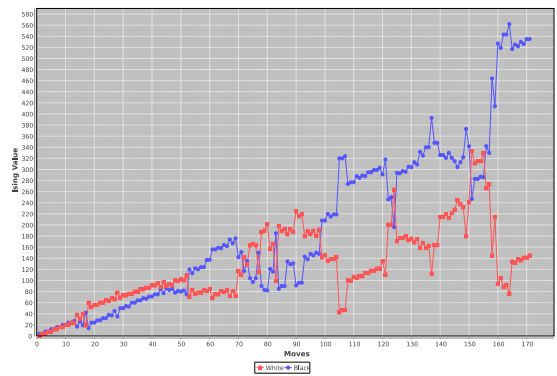

(e) AlphaGo vs. Cho Hanseung 2017

**Fig. 6** Top tournament of AlphaGo please see the official results at https://deepmind.com/research/alphago/match-archive/master/.

Notorious feature of AlphaGo gaming is its ability to keep a kind of equilibrium during any game, that avoids the temptation to try an always increasing control of board. For AlphaGo is quite enough to keep the current advantage, so easily give up to certain board area influence being aware of its strength over the whole board and game. This feature, to some extent, it contrasts to the human Go players' behavior trying to get the absolute prevail over adversary. Human Goers may usually play to get this absolute, and by this mind lost details of weak positions, that eventually lead for loosing points even the whole game. Obsession for ever increasing board control is not a weakness of AlphaGo as a remarkable quality. This mentioned AlphaGo feature requires a well-known knowledge of the game board configuration at any play done.

### 3.3 AlphaGo vs. Lee Sedol games

Lee Sedol is one top Go player who won 18 Go world titles since he was 12 years. AlphaGo – Lee Sedol was a five-game match gamed by March 2016 in Seoul, Korea. Lee Sedol played blacks the odd games and whites the pair games. There, AlphaGo won the first, second, third and fifth game and Lee Sedol the fourth, so 4/5 games won AlphaGo, being the first time a computer defeat a top master so categorical. Figures 7 shows the graphs of each of the five games: G1, G2, G4 and G5, each score similar to the official result. In G3 there is a small difference between the official result and the obtained from simulation.

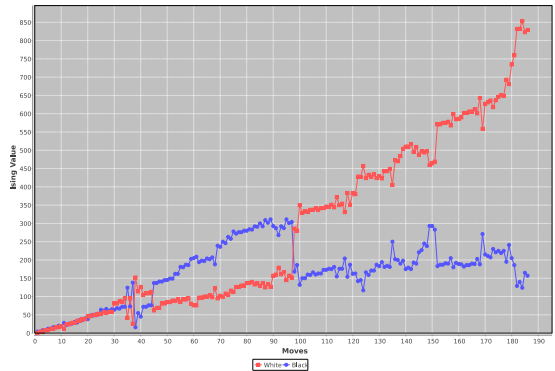
(a) Lee Sedol vs. AlphaGo (G1)

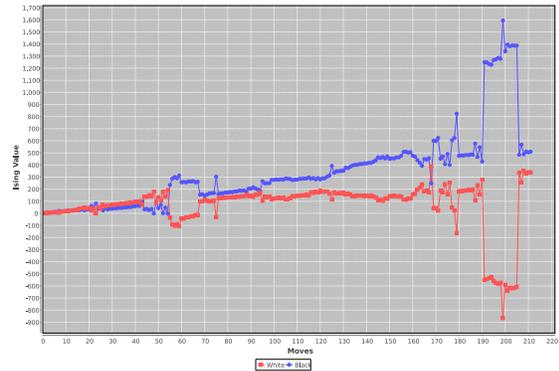
(b) AlphaGo vs. Lee Sedol (G2)

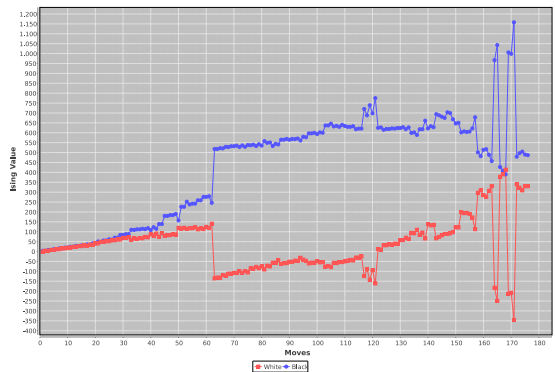
(c) Lee Sedol vs. AlphaGo (G3)

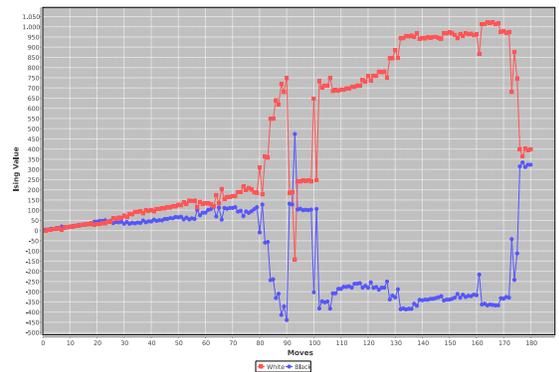
(d) Lee Sedol vs. AlphaGo (G4)

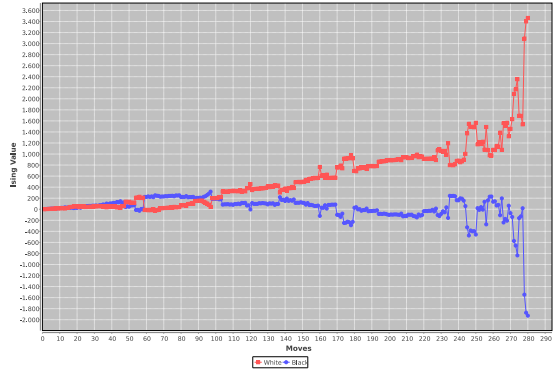

(e) Lee Sedol vs. AlphaGo (G5)

**Fig. 7** Lee Sedol vs AlphaGo https://deepmind.com/research/alphago/match-archive/alphago-games-english/.

## 4. Discussion

Analysis of Go gaming shows the relevance of tactical moves causally connected [35]. Go gaming records on long-term influence moves [15], so the right playing the early moves has strong effects to the late moves outcome options. Historically, Go matches in ancient cultures spend months and most of this time during the first 10 – 15 moves a game in average. Current competition long hours, but most of this time is spent in the first dozen of moves as well.

Go gaming interaction processes among single black and white stones, simple elements –likewise occurring in nature and social major phenomenon– that evolves to complex stones shapes struggling to control most of Go board area. Ising model captures crucial features of interaction phenomena, and we show that quite models the evolution of a Go game as the stochastic process between dichotomy variables, spin black or white assigned to atomic and molecule Go stones; eventually Go gaming interaction displays a phase-transition-like process, when circumstance of balanced black-white dominance abruptly changes, and black stones strength dominance over a given board area overriding white, or conversely.

4.1 Phenomenology of Go game

The closeness between our algorithmic simulation results and data from matches in top Go tournaments make relevant our claim: phenomenology of Go gaming is well comprehended by applying Ising model to analyze the evolution of complex patterns interaction during a game, particularly the phase transitions process occurring there.

The alive and dead stones dynamic is a

The major advance in Go automation it backs in a bio-inspired method trying to emulate the visual cortex of felines and eagles, characterized by an acute vision system. It results in the artificial deep neural network (DNN) [42] that embraces dozens of layers each with long number of neurons; the intense correlation and composition, like in the mentioned animals' visual cortex process, makes DNN a skillfully tool for recognition in complex scenarios. AlphaGo DNN uses convolution integral functions for neurons activation [42].

Even the powerful computer AlphaGo empowerment, the phenomenology of Go gaming is not revealing yet. We claim that phenomenology of Go gaming is superior comprehended modeling Go interaction by means of Ising model. The likely stochastic process of pattern on Go board, constructed by Go tactics during a beat is precisely modeled by Ising Hamiltonian. Some of the interaction dynamic in Go gaming converges to phase transitions, likewise interaction phenomena in physics, chemistry and biology, all they being modeled by Ising model. The stronger interaction among same color stones the weaker interaction among adversarial ones. The energy function modeling each of these processes embraces the atomic or molecular stones each other influencing through time passing and temperature changes. Go gaming temperature corresponds to the player's talent displayed during the game.

We observe that the end behave of Ising Hamiltonian for Go not fits the usual thermodynamics behave. At the end of any game there is ever one winner, so, the global systems is out of the thermodynamics equilibrium.

4.2 Ising model in nature and social sciences

Classical Ising model deal with interaction and transitions properties occurring in statistical physics [31], chemistry phenomena [24], and population biology [25]; recently, Ising model apply is clever in nuclear medicine imaging [31], neural networks [32] and kinetics of protein aggregation studies [34], and in complex pattern recognition on biological processes to classify molecular or tissues patterns, by managing huge databases in Bioinformatics [36-38]. In economy Ising model is applied to analyze non-equilibrium phase transitions [26] or macroeconomic modeling, where emergence of patterns resulting from the collective interaction of a multitude of elementary components or agents is observed [33]. In decision making approach a phase transition threshold, with spontaneous symmetry-breaking of prices, leads to spontaneous valuation in absence of earnings [26], similar to emergence of spontaneous magnetization in absence of a magnetic field. In game theory, the interaction of the spin-player and the state-action in the Ising model energy function is for payoff then apply the Nash equilibrium [27]; as well to analyze repeated games [28], or to modeling systems on next nearest neighbor interaction with phase transitions [29]. In Evolutionary Dynamics of cooperation in the Prisoner's Dilemma game, results on a dipole-model-like, interpreting the start of a lattice of cooperation as a thermo dynamical phase transition [30].

## 5. Conclusions

Phenomenology of Go gaming as dichotomy variables interaction process is clarified by Ising model. Experimental computer simulations allow conclusions that black – white stones' interaction during Go gaming is traced by means of the proposed Ising model energy function. Strength of any Go stones pattern is precisely calculated as a result of the relative positions among ally stones as well as with respect to adversaries that dynamically change during the match evolving. Evolution in Go gaming patterns eventually yields to phase-transition-like phenomena occurring when one stone placement strength territory control at some board area that overrides adversaries. During a Go match any stone at board is affected from the global board state, which effect may be seen as the external field in the Ising model.

**Acknowledgment:** A. Yee special thanks to PROFAPI *Programa de Fomento y Apoyo a Proyectos de Investigación*, with number **PROFAPI2015/ 304.** All the authors thank to Dr. Ricardo Quintero Zazueta, the top senior Mexican Go player and researcher of the Education Mathematics Department in CINVESTAV. He suggested the two classical Go games we simulate and makes suggestions to subtle points in the paper.